\documentclass[pdflatex,sn-mathphys]{sn-jnl}

\usepackage{natbib}
\usepackage{aliascnt}

\usepackage{textgreek}
\usepackage{caption}
\usepackage{hyperref}
\usepackage{siunitx}
\usepackage{adjustbox}
\usepackage{float}
\restylefloat{table}

\usepackage{graphicx}%

\usepackage{soul,color}
\usepackage{amsmath,amssymb,amsfonts}%
\usepackage{amsthm}%
\usepackage{mathrsfs}%
\usepackage[title]{appendix}%
\usepackage{xcolor}%
\usepackage{textcomp}%
\usepackage{manyfoot}%
\usepackage{booktabs}%
\usepackage{algorithm}%
\usepackage{algorithmicx}%
\usepackage{algpseudocode}%
\usepackage{listings}%

\usepackage{multirow}
\usepackage{multicol}

\usepackage{threeparttable}
\usepackage{url}            
\usepackage{nicefrac}       
\usepackage{microtype}      
\usepackage{cleveref}

\newaliascnt{eqfloat}{equation}
\newfloat{eqfloat}{h}{eqflts}
\floatname{eqfloat}{Equation}
\newcommand*{\ORGeqfloat}{}
\let\ORGeqfloat\eqfloat
\def\eqfloat{%
  \let\ORIGINALcaption\caption
  \def\caption{%
    \addtocounter{equation}{-1}%
    \ORIGINALcaption
  }%
  \ORGeqfloat
}

\jyear{2023}%

\begin{document}

\title[Dealing with zero-inflated data: achieving SOTA with a two-fold approach]{Dealing with zero-inflated data: achieving SOTA with a two-fold machine learning approach}


\author*[1]{\fnm{Jo\v{z}e M.} \sur{Ro\v{z}anec}}\email{joze.rozanec@ijs.si}
\author[1]{\fnm{Ga\v{s}per} \sur{Petelin}}
\author[1]{\fnm{Jo\~{a}o} \sur{Costa}}
\author[1]{\fnm{Bla\v{z}} \sur{Bertalani\v{c}}}
\author[1]{\fnm{Gregor} \sur{Cerar}}
\author[2]{\fnm{Marko} \sur{Gu\v{c}ek}}
\author[1]{\fnm{Gregor} \sur{Papa}}
\author[1]{\fnm{Dunja} \sur{Mladeni\'{c}}}

\affil*[1]{\orgname{Jo\v{z}ef Stefan Institute}, \orgaddress{\country{Slovenia}}}
\affil[2]{\orgname{GoOpti}, \orgaddress{\country{Slovenia}}}


\abstract{In many cases, a machine learning model must learn to correctly predict a few data points with particular values of interest in a broader range of data where many target values are zero. Zero-inflated data can be found in diverse scenarios, such as lumpy and intermittent demands, power consumption for home appliances being turned on and off, impurities measurement in distillation processes, and even airport shuttle demand prediction. The presence of zeroes affects the models' learning and may result in poor performance. Furthermore, zeroes also distort the metrics used to compute the model's prediction quality. This paper showcases two real-world use cases (home appliances classification and airport shuttle demand prediction) where a hierarchical model applied in the context of zero-inflated data leads to excellent results. In particular, for home appliances classification, the weighted average of Precision, Recall, F1, and AUC ROC was increased by 27\%, 34\%, 49\%, and 27\%, respectively. Furthermore, it is estimated that the proposed approach is also four times more energy efficient than the SOTA approach against which it was compared to. Two-fold models performed best in all cases when predicting airport shuttle demand, and the difference against other models has been proven to be statistically significant. 
}




\maketitle

\section{Introduction}
Supervised machine learning models exploit relationships between a dependent variable and a set of independent variables encoded in a given dataset. Such models can help explore and understand relationships in data and reduce such relationships to a few parameters. Nevertheless, datasets with abundant zeros in the dependent variable (zero-inflated datasets) are common in certain domains. The abundance of zeroes affects the learning process of machine learning models and renders them incapable of properly predicting zero and non-zero values \cite{abraham2009semi}. Furthermore, the abundance of zeroes distorts the metrics, making it difficult to assess the quality of the model's predictions \cite{rovzanec2022reframing}.

To deal with zero-inflated regression data, multiple approaches have been tried. Such approaches have been classified into zero-truncated models \cite{cohen1954estimation} (assume the data comes from a single underlying distribution, where negative values are censored and stacked on zero; e.g., the Tobit model \cite{tobin1958estimation}) and hurdle models \cite{cragg1971some,mullahy1986specification} (two-component mixture models, where one component is concerned with modeling zeroes, while the second one models non-zero occurrences). Nevertheless, Lambert \cite{lambert1992zero} is recognized among pioneering authors regarding hurdle models, being among the first ones who provided a throughout analysis of zero-inflated models in the presence of covariates and proposed a zero-inflated Poisson regression model to predict defects in a manufacturing setting by modeling a perfect state (where no faults are expected) and an imperfect state (where faults are expected). Since then, many works proposed enhanced approaches. While such an approach has been used and developed in statistics, its application has been scarce when considering machine learning \cite{abraham2009semi,rovzanec2022reframing}. Furthermore, failure to frame zero-inflated regression problems as a two-fold approach has also affected performance measurement for such models. Finally, while hierarchical approaches have been documented in the literature to address multiple classification problems related to sensor data (e.g., to classify human locomotion \cite{narayan2021real}), we have not found household classification problems framed this way.

Given the two-fold approach in hurdle models, \cite{abraham2009semi} measured performance considering the accuracy and F-measure to determine classification performance (zero vs. non-zero occurrences) and RMSE to measure the regression performance (including zeroes). Nevertheless, the accuracy metric is inadequate for classification performance in imbalanced datasets. Furthermore, the F-measure is constrained to an arbitrary cut-off threshold. To address this issue, \cite{rovzanec2022reframing} proposed measuring classifier performance considering the Area Under the Receiver Operating Characteristic Curve (AUC ROC, see \citep{BRADLEY19971145}). When measuring the quality of the regression models, RMSE has the desired property of penalizing large errors (e.g., when a non-zero value should have been predicted but was not). Nevertheless, \cite{rovzanec2022reframing} proposed measuring regressors' performance with two variants of the MASE metric \cite{hyndman2006another}. Computing the MASE metrics only on a subset of the test data guarantees that regression performance is measured regardless of zero-inflated data and avoids measurement distortions. Furthermore, the regressor's quality assessment benefits from other MASE metric properties, such as scale invariance, predictable behavior when predicted values are close or equal to zero, symmetry, interpretability, and asymptotic normality. When performing classification in a zero-inflated context, we consider the quality of the classifier is assessed twice with the AUC ROC metric: (i) to measure the first classifiers' capability to predict whether a zero or non-zero value is expected and (ii) to measure the second classifiers' capability to predict a particular class of interest.

\begin{figure}[ht!]
\centering
\includegraphics[width=0.48\textwidth]{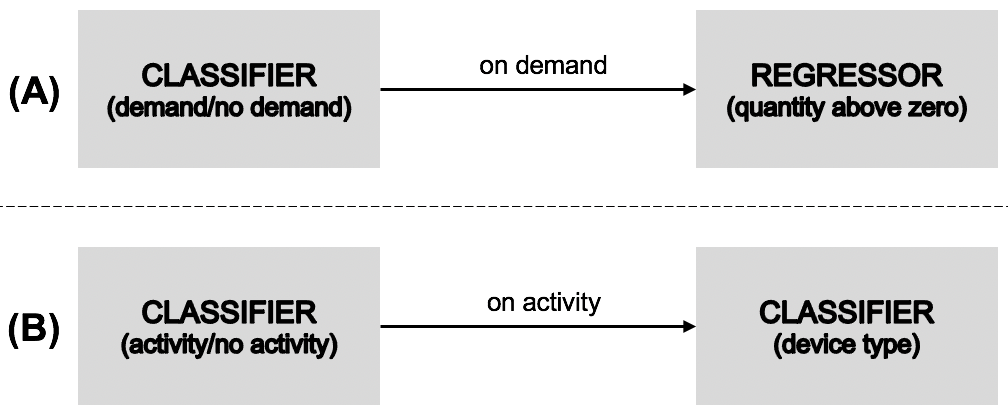}
\caption{(A) depicts how a two-fold model is used to predict the airport shuttle passengers' demand: the classifier determines whether some demand will exist, and the regressor predicts the expected number of passengers. (B) depicts how a two-fold model is used to predict the home appliance type based on sensor measurements: the first classifier determines whether some appliance is being used. In contrast, the second one determines the appliance type.}
\label{F:TWO-FOLD-MODELS}
\end{figure}

The paper describes two real-world use cases where machine learning models are applied to zero-inflated datasets and where these outperform existing SOTA regression models. While the models are different in each case (features are domain-specific, and different models are of interest for each use case), the same approach is used in both cases. In particular, we consider two-fold regression models when predicting airport shuttle passengers' demand and two-fold classifiers to classify home appliances based on their energy consumption (see Figure \ref{F:TWO-FOLD-MODELS} for details).

This paper is organized as follows. First, we describe the use cases where the abovementioned approach is applied. To the extent of our knowledge, two-fold machine learning models in zero-inflated datasets have not been applied yet to the domains represented by the use cases we tackled. Furthermore, the section introduces the datasets used for each use case. Second, we describe the experiments performed and later present the results obtained for each use case. Finally, we present the conclusions and outline future work.


\section{Use cases}\label{S:USE-CASES}

\subsection{Home appliances classification}\label{SS:UC-HOME-APPLIANCES}
Environmental concerns are heightened globally due to the rising energy consumption associated with population growth and technological advancements. Because of that, next-generation smart grids allow bidirectional transmission of power and data, creating an automated power grid~\cite{FangSG} is seen as one of the enablers of increased efficiency. Part of the enabling technologies is also monitoring active home appliances in smart grids. Appliance classification plays a crucial role in smart grid demand response scenarios, which involve adjusting energy consumption based on supply and demand conditions. By providing real-time data on the energy usage of different appliances, the information helps implement demand response strategies by selectively controlling or prioritizing specific devices during peak demand periods or when renewable energy sources are limited. Additionally, accurate appliance classification data can facilitate more precise billing and tariff structures. This enables the implementation of time-of-use tariffs, where different energy prices apply during peak and off-peak hours. Appliance classification helps in accurately attributing energy usage of individual appliances to provide detailed consumption information to consumers and determining fair billing based on each appliance consumption.

Due to the nature of specific home appliances, like toasters, washing machines, and others, a sparse operation cycle can be observed for such devices. For example, a person might use the toaster in the morning to make breakfast, but it would not be used for the rest of the day. Additionally, constant deep learning model inference can result in a higher energy consumption\cite{cerar2022resource}, compared to the energy savings enabled by the technology. 

\begin{figure}[ht!]
\centering
\includegraphics[width=0.48\textwidth]{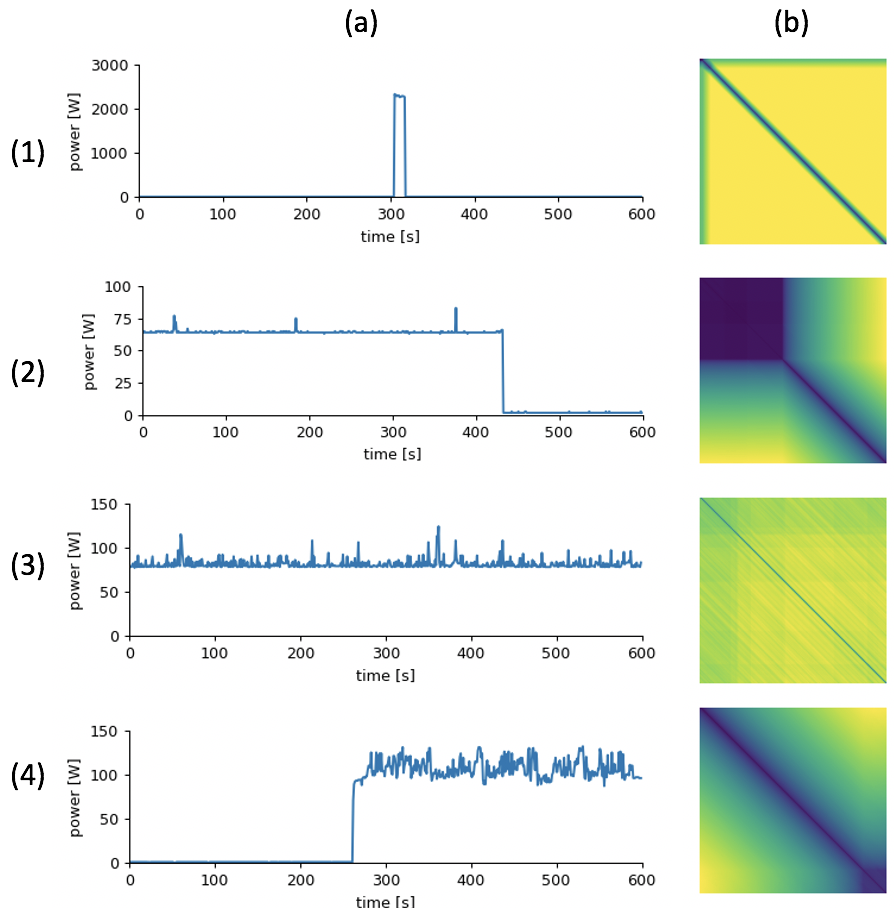}
\caption{The plots represent boiler samples from the UKDALE dataset. The samples refer to the (1) HEKA, (2) HTPC, (3) TV, and (4) home server (NAS) appliances. Column (a) shows the time series reporting power consumption, while column (b) displays recurrence plots capturing the time series information. The recurrence plots are used as an input to the VGG11 model.}
\label{F:APPLIANCE-TIME-SERIES-AND-PLOTS}
\end{figure}

\begin{figure}[ht!]
\centering
\includegraphics[width=0.48\textwidth]{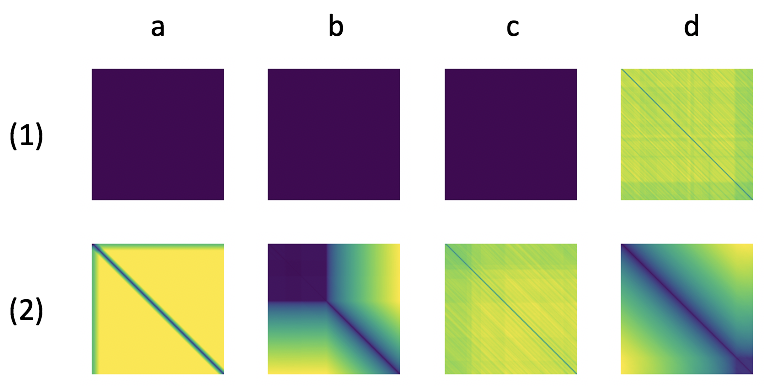}
\caption{The recurrence plots represent different appliances from the UKDALE dataset: (a) HEKA, (b) HTPC, (c) TV, and (d) home server (NAS). The recurrence plots at the top take into account the whole measurement period, while those at the bottom were computed considering time series segments related to activity periods only.}
\label{F:APPLIANCE-RECURRENCE-PLOTS}
\end{figure}

This use case demonstrates on a real-world dataset how a simple two-fold machine learning approach can enhance the performance of the classification of home appliances in next-generation smart grids. 75\% of the sensor measurements correspond to device inactivity, and therefore, just 25\% are left to learn characteristic patterns that enable discriminating between devices. Figure \ref{F:APPLIANCE-TIME-SERIES-AND-PLOTS} shows how recurrence plots change when only activity periods are extracted. Figure \ref{F:APPLIANCE-RECURRENCE-PLOTS} shows recurrence plots (whole measurements vs. activity only) for five device types. It can be observed that recurrence plots where only periods of activity are considered are substantially different from each other vs. those that consider the whole measurement period.

\subsection{Shuttle transfer demand}

Shuttle transfers provide convenient and efficient transportation to individuals or groups of passengers between airports, hotels, and other designated locations. Passengers are pooled in vehicles based on their departure time, flexibility, and routing preferences, resulting in efficient and cost-effective transportation solutions. To optimize sales and operations, it is crucial to have advanced knowledge of the number of passengers. This information offers two significant advantages: enabling more efficient vehicle allocation and facilitating dynamic pricing strategies. For instance, having information about the expected number of passengers booking a particular route at a specific time allows for precise vehicle allocation based on matching the capacity requirements. In addition, when the number of passengers that will travel together on a particular day and hour is known during the booking process, it enables a fairer distribution of the costs associated with the ride. Unfortunately, long-term prediction of the number of passengers is relatively challenging due to its susceptibility to factors like holidays, plane departures and arrivals, etc. Furthermore, passengers often book their rides months, weeks, or even just a few days in advance, making it difficult to allocate enough vehicles on time and offer a fair fixed price to early bookers without knowing how many people will be pooled together in a single vehicle.

The paper focuses on a real-world shuttle company for which three specific routes (A$\rightarrow$B, A$\rightarrow$C, and A$\rightarrow$D) were selected. In these routes, the shuttles transport passengers between 15\% and 50\% of the time (leading to a sparsity between 50\% and 85\%). Accurately predicting the number of passengers traveling between two cities and the optimal timing is critical to appropriate planning. This involves predicting the occurrence of at least one reservation (i.e., whether a vehicle will be required to travel) and the number of passengers to be transported (i.e., the number of reservations). Due to these two reasons, the problem is a natural fit for a two-fold regression.

\section{Experiments}\label{S:EXPERIMENTS}

\subsection{Home appliances classification}\label{SS:EXPERIMENTS-UC-HOME-APPLIANCES}
This research contrasts the usual SOTA approach to home appliance classification with a two-fold classification approach. The two-fold machine learning models consist of (a) a classification model to predict home appliance activity and (b) a classification model to discriminate the kind of appliance used. While the first classifier is trained as a binary classifier over all the data (considering two cases: activity or no activity), the second classifier is trained only on time series segments displaying appliance activity.

\textbf{Dataset:} 
The UK-DALE~\cite{kelly2015uk} dataset was used, which has real-world measurements of power consumption from 54 appliances collected over 655 days in five households in the United Kingdom. Measurements were collected with a sampling time of six seconds, ideally resulting in 600 measurements per one-hour time window. All time series samples from the dataset were divided into sub-intervals with a one-hour fixed window length, and samples with missing data were discarded. Appliances that had less than ten percent of samples of the largest class were also discarded. Using domain knowledge, fridges and freezers were merged into a single category (i.e., ``fridge/freezer''), and the same was done for toasters and kettles (i.e., ``HEKA''), given they display similar patterns. As a result, twelve different appliance classes were considered for the experiments. The final dataset contains $1\,275\,508$ samples, where 25\% ($319\,830$) samples contain actual activity.

\textbf{Features:} To determine whether some home appliance is active, the raw measurements of the last time hour interval were considered. To determine which appliance is active, time series were transformed into images leveraging recurrence plots \cite{eckmann1995recurrence}. A deep learning architecture was used to extract features and determine the type of appliance based on such images. 

\textbf{Machine learning models:} Two different machine learning models were considered. First, a VGG11 \cite{simonyan2014very} architecture was used to discriminate between home appliances based on recurrence plots. This combination has already shown SOTA results in other domains~\cite{Bertalanic_9715175}. The study of idle states of devices has shown that some devices, even inactive, can still consume some power when idle. Given the range of idle power consumption varies between devices, a threshold method cannot correctly detect idle or active states for an extensive range of devices. Therefore, an XGBoost classifier was trained to predict whether an appliance was active during the observation window. The appliance type was determined with a VGG11 model, leveraging a recurrence plot built considering only the appliance activity period. The classes were weighted during the training procedure to mitigate class imbalance.

The XGBoost classifier was used with default hyperparameters (XGBoost v1.7.6 implementation). The VGG11 architecture was trained from scratch using Adam optimizer with a learning rate of $0.001$, batch size $128$, and cross-entropy as a loss metric. The training duration was limited to 20 epochs.

The models were trained with a stratified K-Fold shuffle-and-split approach (80\% assigned to the train set and 20\% to the test set). Due to the long training time, the experiment was performed only once. 

\textbf{Metrics:} The models' performance was measured with four metrics: F1, precision, recall (the three of them reported in previous works - we report them to provide comparability), and the AUC ROC, which summarizes the overall models' discriminative power while not being affected by the class imbalance. The AUC ROC is reported twice for the two-fold model: once to measure how accurately the model discriminates between activity vs. non-activity periods and once to assess the overall discriminative capability when considering appliance types.

\textbf{Hardware:} All tests were conducted using a computer equipped with two AMD(R) EPYC(R) 75F3 CPUs operating at 3GHz, 1TB of RAM, 3g.40gb instance of NVIDIA(R) A100, and running Ubuntu operating system.

\textbf{Model complexity and theoretical energy consumption:} According to the guidelines considered in ~\cite{Bertalanic_9715175}, the complexity of the model was measured in floating point operations (FLOPs) needed for prediction. Based on the calculated FLOPs and the provided hardware details, a theoretical energy consumption per prediction (TEC) can be calculated. TEC was calculated by the equation $TEC = \frac{FLOPs}{ FLOPS /Watt}$ where $FLOPS/Watt$ represents floating point operations per second per watt. Theoretical $FLOPS/Watt$ for an A100 Nvidia GPU FP64 precision is 38.8 $GFLOPS/Watt$.

\subsection{Shuttle transfer demand}\label{SS:EXPERIMENTS-SHUTTLE-TRANSFER}
This research contrasts the usual SOTA approach to home appliance classification with a two-fold classification approach. The two-fold machine learning models consist of (a) a classification model to predict home appliance activity and (b) a classification model to discriminate the kind of appliance used. While the first classifier is trained as a binary classifier over all the data (considering two cases: activity or no activity), the second classifier is trained only on time series segments displaying appliance activity. 

Regression models that achieved SOTA performance in a particular dataset against two-fold machine learning models inspired by zero-inflated regression approaches. The two-fold machine learning models consist of (a) a classification model to predict when zero or non-zero values occur and (b) a regression model to predict the non-zero values. While the classification model is trained as a binary classifier over all the data (considering two possible classes: zero or non-zero), the regressor is trained only on non-zero data to avoid issues derived from zeroes occurrence. The following comparisons are made in all cases: (a) zero-inflated regression models vs. traditional regressors that achieved SOTA performance on such datasets and (b) local vs. global models for classification and regression. While local models are trained only on the data of a particular destination, global models are trained on the data of the three destinations under consideration. 

\textbf{Dataset:} 
The dataset comprises hourly demand data collected over four years. Due to its relatively low demand density, the data is grouped into intervals of three hours. As a result, when making demand forecasts, the passengers count departing from a particular city to some destination is anticipated within a three-hour timeframe.

\textbf{Features:}
A set of 14 hand-crafted features were created. A feature was created to indicate whether the day corresponded to a holiday. Temporal attributes such as hour, day of the month, month, and day of the week were considered. Cyclical timestamp features were used to capture cyclic temporal patterns, effectively encoding recurring patterns within intervals such as quarter days, half days, full days, and weeks. Furthermore, a variable was used to discriminate between weekdays and weekends.

All features were scaled using a standard scaler which substracts the mean and scales values to unit variance. All the regressors had their targets scaled using the $\log(1 + x)$ transformation before training. After making a prediction, its value was rescaled to its original values using the $e^x - 1$ transformation.

\textbf{Machine learning models:} The experiments considered three regressor machine learning algorithms trained over all of the time series values: A Histogram-Based Gradient Boosting (HGB) tree regressor, a Multi-Layer Perceptron (MLP), and a Linear Regression (LR). The LR was considered due to its simplicity. The two-fold models considered a Histogram-Based Gradient Boosting (HGB) tree classifier combined with a machine learning regressor (either LR, MLP, or Support Vector Regressor (SVR)). While the classifier was trained on the whole time series, the regressors were trained only on non-zero data. The Youden index \cite{youden1950index} (defined as $J=sensitivity+specificity-1$) was computed to determine the best threshold cut-off for every classification model. Given the high sparsity levels in the data, a zero predictor was considered a baseline model. The default hyperparameters provided by \textit{scikit-learn} version \textit{1.3.0} were used in the experiments. For the MLP regression task, a model with a single hidden layer comprising 100 neurons, a ReLU activation function, and an Adam optimizer was used. A model with a radial basis function kernel and a regularization set to 1.0 was employed for the SVR analysis. With a histogram-based gradient boosting tree, a learning rate of 0.1 was used with a maximum iteration limit of 100. The log loss and squared error loss were applied in classification and regression models, respectively.

A time series cross-validation setting was considered, forecasting 24 hours ahead. The models' performance was tested in the last month contained in the dataset. For each new day considered in the test set, the models were retrained with the information available up to 24 hours before the predicted time slot. The experiments were repeated twenty times to ensure the results were reliable and accounted for the stochasticity of some machine learning algorithms.

\textbf{Metrics:}
Two metrics were considered for measuring the models' performance: AUC ROC (to measure how well the models predict when a demand event will occur) and MASE (to measure how well the model estimates the demand quantity). Furthermore, Wilcoxon paired rank tests were performed to determine the statistical significance of the results. 

While two-fold models provide classifier predictions for assessing the AUC ROC, this is not true for usual regressors. Therefore, three strategies were devised to turn regression predictions into predictions signaling event occurrence. First, consider any value below one a non-demand event, and every value equal or greater than one to signal a demand event (AUC ROC\textsubscript{floor}). Second, consider any value equal to or below 0.5 to be a non-demand event and every value above that to be a demand event (AUC ROC\textsubscript{round}). Finally, consider only zero values as non-demand events and any value above zero as a demand event (AUC ROC\textsubscript{ceil}).

MASE was computed with the following equation: 

Following \cite{rovzanec2022reframing}, two MASE variants were considered (namely MASE\textsubscript{I} and MASE\textsubscript{II}). MASE\textsubscript{I} was computed on the test data that results from ignoring zero values. On the other hand, MASE\textsubscript{II} was computed considering all points where either a non-zero value was observed, or the classification model predicted non-zero occurrence. While MASE\textsubscript{I} describes the model's performance when true events occur, MASE\textsubscript{II} provides insights into the cost of false positives. When the model predicting event occurrence performs perfectly, MASE\textsubscript{I} equals MASE\textsubscript{II}.

\textbf{Hardware:} All tests were conducted using a computer equipped with an Intel(R) Xeon(R) CPU E5-2680 v3 operating at 2.50GHz, 1 TB of RAM, and running the Ubuntu operating system.

\section{Results and Discussion}\label{S:RESULTS-AND-DISCUSSION}

\subsection{Home appliances classification}\label{SS:RESULTS-HOME-APPLIANCE}

Table \ref{T:RESULTS-APPLIANCES} presents the home appliance classification results. At the two-fold models, the XGBoost classifier predicting whether some home appliance activity is detected achieved an AUC ROC of 0.9901. The home appliance classification outcomes show that using a two-fold approach enhanced the overall model's discriminative performance by AUC ROC 0.2119 points on average; and up to 0.2653 AUC ROC points in the best case, leading to almost perfect classification performance in many cases. The rest of the metrics mirrored this improvement. In particular, the weighted average between both approaches showed Precision, Recall, and F1 increased by 0.1858, 0.2209, and 0.2866 points, respectively. The cases where Precision increased the most were TV detection (0.2196 points increase) and HEKA (0.1977 points increase). Recall improvements were notorious. E.g., HEKA increased recall from 0.2117 to 0.8990 (0.6873 points), the computer monitor Recall increased from 0.1077 to 0.7016 (0.5939 points), and the laptop computer from 0.1614 to 0.7808 (0.6194 points). Substantial Precision and Recall improvements were also mirrored in F1 scores. E.g., the washer dryer improved the F1 score by 0.5638 points, the desktop computer by 0.5385 points, and the computer monitor by 0.5247 points.

On top of the SOTA results achieved by the two-fold approach, we expect it to show a lower energy consumption, given the deep learning model (more resource intensive than the XGBoost) is utilized only when some appliance activity is predicted. Following the method proposed by ~\cite{Bertalanic_9715175}, the model approach requires about 13.3 GFLOPs per prediction. The dataset used consisted of 1275508 samples, of which 319830 were labeled with activity. Without the two-step classification process, TEC would consume $\approx437.2$ kJ of energy for all 1275508 samples, compared with $\approx109.6$ kJ for activity classes only. This means that about four times less energy is consumed, making the approach proposed in this paper much more energy efficient considering that the computational complexity of XGBoost is negligible compared to the deep learning approaches.

\begin{table*}[!htbp]
	\centering
	\footnotesize
	\begin{threeparttable}[b]
		\caption{UKDALE classification results. In the last column, AUC uses a one-vs-rest approach and weighted average. \label{T:RESULTS-APPLIANCES}}
		\label{tab:results-classification-ukdale}
		\begin{tabular}{@{}lllllllllllll|ll@{}}
			\toprule
			
			\multirow{2}{*}{Class}
			& \multicolumn{4}{c}{Sparse time series data}
			& \phantom{}
			& \multicolumn{4}{c}{Time series data}
            & \phantom{}
			\\\cmidrule{2-5}\cmidrule{7-10}
			
			& Precision
			& Recall
			& F1
            & AUC ROC
			& 
			& Precision
			& Recall
			& F1
            & AUC ROC
            & 
			\\\midrule

                HEKA     & 0.7153    & 0.2117    & 0.3267 & 0.7396 &&
                            0.9139   & 0.8990    & 0.9064 & 0.9943 \\   
      fridge/freezer     & 0.9376    & 0.9171    & 0.9272 & 0.9803 && 
                            0.9761   & 0.9831    & 0.9796 & 0.9991 \\    
                HTPC     & 0.7606    & 0.3859    & 0.5120 & 0.9217 && 
                            0.8064   & 0.8692    & 0.8366 & 0.9834 \\
              boiler     & 0.8476    & 0.7069    & 0.7709 & 0.9377 && 
                            0.8549   & 0.8810    & 0.8678 & 0.9881 \\     
    computer monitor     & 0.6981    & 0.1094    & 0.1891 & 0.7319 && 
                            0.7016   & 0.7265    & 0.7138 & 0.9833 \\     
    desktop computer     & 0.7722    & 0.1077    & 0.1891 & 0.7324 && 
                            0.8088   & 0.6612    & 0.7276 & 0.9841 \\     
     laptop computer     & 0.7568    & 0.1614    & 0.2661 & 0.7423 && 
                            0.7953   & 0.7808    & 0.7880 & 0.9862 \\     
               light     & 0.6149    & 0.9651    & 0.7512 & 0.7363 && 
                            0.8806   & 0.8696    & 0.8750 & 0.9784 \\    
           microwave     & 0.6761    & 0.2065    & 0.3164 & 0.7748 && 
                            0.8489   & 0.8344    & 0.8416 & 0.9956 \\     
     server computer     & 0.6454    & 0.8136    & 0.7198 & 0.9812 && 
                            0.7513   & 0.8803    & 0.8107 & 0.9965 \\       
          television     & 0.5396    & 0.3672    & 0.4370 & 0.9327 && 
                            0.7592   & 0.7486    & 0.7538 & 0.9787 \\     
        washer dryer     & 0.8865    & 0.0862    & 0.1572 & 0.7082 && 
                            0.9460   & 0.5825    & 0.7210 & 0.9735 \\     
			
			\midrule
			Weighted avg. & 0.6835   & 0.6470   & 0.5805 & 0.7745
            && 0.8693    & 0.8679    & 0.8671 & 0.9864\\
			\bottomrule
		\end{tabular}	
	\end{threeparttable}
\end{table*}

\subsection{Shuttle transfer demand}\label{SS:RESULTS-SHUTTLE-TRANSFER}
The performance for the best and second-best models, along with the zero predictor model and best-performing regressor (that is not a two-fold model), are presented in Table \ref{T:RESULTS-GOOPTI}. In all cases, the two-fold regression models led to the best results. In particular, we observed that the best results were achieved by the local two-fold models and followed by the global two-fold models. In most cases, using data from other Source$\rightarrow$Destination paths (global models) negatively affected the predictor outcomes. An interesting finding about the regressive power of the models is that most of them while having predictive power about when a demand event will take place, had little predictive power regarding the number of passengers that will require the service. This is evident from the MASE\textsubscript{I} values, which are usually above one. Therefore, two conclusions can be drawn: (i) usually, better results can be obtained by just predicting the average number of passengers as learned from the train set, and (ii) additional features are required to signal better the number of potential passengers for a given route. It must be noted that the SVR regressor trained on top of the classifier for two-fold models performed best in most cases and that the difference between regressors trained at the two-fold models was statistically significant when tested with the Wilcoxon signed-rank test at a $p-value=0.05$ for two routes. Furthermore, the models' performance measured across the different metrics for the models displayed in Table \ref{T:RESULTS-GOOPTI} is considered statistically significant when performing the Wilcoxon signed-rank test at a $p-value=0.05$.

While the results in Table \ref{T:RESULTS-GOOPTI} show the best AUC ROC obtained for each regression model, these could correspond to different class attribution strategies as detailed in Section "Experiments - Shuttle transfer demand". To illustrate this, we provide a few examples in Table \ref{T:RESULTS-SHUTTLE-AUC}. The table shows that different class attribution strategies may result in different performance measurements. E.g., while AUC ROC\textsubscript{round} score for the \textit{LR on all values} model is 0.7079, the same regression predictions led to  AUC ROC\textsubscript{floor} and AUC ROC\textsubscript{ceil} scores of 0.5791 and 0.5000. While the ceiling of the values removed any intuition about when the demand will take place, rounding them strengthened the signal and led to competitive results.

\begin{table*}[ht!]
\centering
\caption{Best and second-best local and global models and the zero predictor model are displayed. The best results are bolded, and the second-best are underlined. \label{T:RESULTS-GOOPTI}}
\resizebox{0.80\textwidth}{!}{
\begin{tabular}{@{}llllll@{}}
\toprule
Source$\rightarrow$Destination & Type & Model & Best AUC ROC & MASE\textsubscript{I} & MASE\textsubscript{II} \\
\midrule
\multirow{7}{*}{A$\rightarrow$B} & ---                          & Zero Predictor                        & 0.5000               & 0.6875                & 0.6875 \\
\cmidrule{2-6}
                                 & \multirow{4}{*}{local}       & \textbf{Classifier (HGB) + LR}        & \textbf{0.6908}      & \textbf{1.3766}       & \textbf{1.4139} \\
                                 &                              & \textbf{Classifier (HGB) + MLP}       & \textbf{0.6908}      & \textbf{1.3766}       & \textbf{1.4139} \\
                                 &                              & \textbf{Classifier (HGB) + SVR}       & \textbf{0.6908}      & \textbf{1.3766}       & \textbf{1.4139} \\
                                 &                              & HGB Regression on all values          & 0.5066               & 0.9338                & 0.9512 \\
\cmidrule{2-6}
                                 & \multirow{2}{*}{global}      & \underline{Classifier (HGB) + LR}     & \underline{0.5903}   & \underline{1.3375}    & \underline{1.3375} \\
                                 &                              & HGB Regression on all values          & 0.5625               & 1.5574                & 3.7406 \\
\midrule
\multirow{5}{*}{A$\rightarrow$C} & ---                          & Zero Predictor                        & 0.5000               & 0.9120                & 0.9120 \\
\cmidrule{2-6}
                                 & \multirow{2}{*}{local}       & \textbf{Classifier (HGB) + SVR}       & \textbf{0.7882}      & \textbf{1.3118}       & \textbf{1.4945} \\
                                 &                              & HGB Regression on all values          & 0.7396               & 0.9303                & 0.9391 \\
\cmidrule{2-6}
                                 & \multirow{2}{*}{global}      & \underline{Classifier (HGB) + SVR}    & \underline{0.7495}   & \underline{0.9469}    & \underline{0.9927} \\
                                 &                              & LR on all values                      & 0.7327               & 0.8468                & 0.8833 \\
\midrule
\multirow{5}{*}{A$\rightarrow$D} & ---                          & Zero Predictor                        & 0.5000               & 0.6883                & 0.6883 \\
\cmidrule{2-6}
                                 & \multirow{2}{*}{local}       & \textbf{Classifier (HGB) + SVR}       & \textbf{0.6931}      & \textbf{1.5852}       & \textbf{1.7654} \\
                                 &                              & HGB Regression on all values          & 0.6736               & 1.1372                & 1.1912 \\
\cmidrule{2-6}
                                 & \multirow{2}{*}{global}      & \underline{Classifier (HGB) + SVR}    & \underline{0.6836}   & \underline{1.0640}    & \underline{1.0976} \\
                                 &                              & HGB Regression on all values          & 0.6670               & 1.0634                & 1.0876 \\
\bottomrule
\end{tabular}}
\end{table*}
\begin{table*}[ht!]
\centering
\caption{An example of AUC ROC values that result from different class attribution strategies. \label{T:RESULTS-SHUTTLE-AUC}}
\resizebox{0.80\textwidth}{!}{
\begin{tabular}{llllll}
\toprule
Source$\rightarrow$Destination & Type & Model & AUC ROC\textsubscript{round} & AUC ROC\textsubscript{floor} & AUC ROC\textsubscript{ceil} \\
\midrule
\multirow{4}{*}{A$\rightarrow$C}& --- & Zero Predictor  & 0.5000 & 0.5000 & 0.5000 \\
\cmidrule{2-6}
                  & \multirow{3}{*}{local} & LR on all values & 0.7079 & 0.5791 & 0.5000 \\
                  &   & MLP on all values  & 0.7189 & 0.5907 & 0.5000 \\
                  &   & HGB Regression on all values & 0.7396 & 0.6785 & 0.5000 \\
\bottomrule
\end{tabular}}
\end{table*}

\section{Conclusions and future work}\label{S:CONCLUSION}
This research addressed the problem of zero-inflated data by considering two-fold machine learning models to solve real-world regression and classification problems. 

The results show that such an approach achieves excellent results when predicting shuttle transfer demand. While regular regression models sometimes learn when certain shuttle transfers will be required, they may fail to do so, resulting in spurious predictions. On the other hand, two-fold models adequately captured the demand event occurrence in all cases. Nevertheless, when predicting the number of passengers per shuttle, the models lagged compared to predicting the average number of passengers observed in the training period (MASE was slightly above 1 in most cases). Therefore better features are required to capture dynamics influencing the number of travelers. 

The two-fold approach achieved SOTA results for home appliance classification. By learning activity and non-activity patterns, the data used to classify appliance types is much more informative, leading to better classification results. In particular, when comparing the weighted average of Precision, Recall, F1, and AUC ROC, we observed performance increased by 27\%, 34\%, 49\%, and 27\%, respectively. Furthermore, it is estimated this approach is four times more energy efficient than the SOTA method it was compared against.

Future work regarding home appliances classification will focus on complementing the two-fold approach with regression models that leverage graph representations of time series to reduce the computational cost and obtain more energy-efficient machine learning models while still achieving SOTA results.

Future work regarding shuttle demand prediction will focus on two lines of work: (a) dataset enrichment and feature engineering (e.g., considering passengers' most frequent travel destinations) to enhance the existing models and (b) building a graph with geographic and temporal data regarding the shuttle routes and apply graph machine learning algorithms to it to forecast demand.

\backmatter

\bmhead{Acknowledgments}
This work was partially funded by the Slovenian Research Agency (ARRS/ARIS) under Grant P2-0016, P2-0098, and young researcher grants. The work is also part of a project funded by the European Union from Horizon Europe under grant agreement No 101077049 (CONDUCTOR).






\bibliography{main}


\end{document}